%% file: main.tex
\documentclass[sigconf,screen]{acmart}
\setcopyright{none}
\renewcommand\footnotetextcopyrightpermission[1]{} 
\AtBeginDocument{%
  }

\usepackage{graphicx}
\usepackage{subcaption}
\usepackage{multirow}
\usepackage{adjustbox}
\usepackage{booktabs}
\usepackage{amsmath}
\begin{document}

\title{What You Perceive Is What You Conceive:  A Cognition-Inspired Framework for Open Vocabulary Image Segmentation}

\author{
    {Jianghang Lin$^{1}$, Yue Hu$^{1}$, Jiangtao Shen$^{1}$, Yunhang Shen$^{2}$, Liujuan Cao$^{1}$, Shengchuan Zhang$^{1}$, Rongrong Ji$^{1}$}\\
    {$^{1}$ Key Laboratory of Multimedia Trusted Perception and Efficient Computing,\\ Ministry of Education of China, Xiamen University, China.}\\
    {$^{2}$ Tencent Youtu Lab, China.}\\
    {\{hunterjlin007, huyue1279, shenjt\}@stu.xmu.edu.cn, shenyunhang01@gmail.com,\\ \{caoliujuan, zsc\_2016, rrji\}@xmu.edu.cn}
}

\input{sec/0_abstract}

\maketitle
\input{sec/1_introduction}
\input{sec/2_related_work}
\input{sec/3_method}
\input{sec/4_experiments}
\input{sec/5_conclusion}

\bibliographystyle{ACM-Reference-Format}
\bibliography{main}


\end{document}

%% file: sec/0_abstract.tex
\begin{abstract}
Open vocabulary image segmentation tackles the challenge of recognizing dynamically adjustable, predefined novel categories at inference time by leveraging vision-language alignment.
However, existing paradigms typically perform class-agnostic region segmentation followed by category matching, which deviates from the human visual system's process of recognizing objects based on semantic concepts, leading to poor alignment between region segmentation and target concepts.
To bridge this gap, we propose a novel Cognition-Inspired Framework for open vocabulary image segmentation that emulates the human visual recognition process: first forming a conceptual understanding of an object, then perceiving its spatial extent.
The framework consists of three core components: 
(1) A Generative Vision-Language Model (G-VLM) that mimics human cognition by generating object concepts to provide semantic guidance for region segmentation.
(2) A Concept-Aware Visual Enhancer Module that fuses textual concept features with global visual representations, enabling adaptive visual perception based on target concepts.
(3) A Cognition-Inspired Decoder that integrates local instance features with G-VLM-provided semantic cues, allowing selective classification over a subset of relevant categories.
Extensive experiments demonstrate that our framework achieves significant improvements, reaching $27.2$ PQ, $17.0$ mAP, and $35.3$ mIoU on A-150. It further attains $56.2$, $28.2$, $15.4$, $59.2$, $18.7$, and $95.8$ mIoU on Cityscapes, Mapillary Vistas, A-847, PC-59, PC-459, and PAS-20, respectively.
In addition, our framework supports vocabulary-free segmentation, offering enhanced flexibility in recognizing unseen categories.
Code will be public.
\end{abstract}

%% file: sec/1_introduction.tex
\section{Introduction}
\label{sec:intro}
Image segmentation~\cite{kirillov2019panoptic,cheng2020panoptic,he2017mask,cheng2021per,cheng2022masked} aims to partition an image into distinct regions, each assigned a unique identification ID.
Significant progress has been made in this area, establishing it as a cornerstone of computer vision.
%
However, conventional segmentation paradigm\cite{kirillov2019panoptic,he2017mask,hu2023you,li2022panoptic,wang2021max} typically rely on linear classifiers trained on a closed set of categories, which limits the model’s ability to generalize to novel scenes containing diverse and semantically rich objects.
%
To address this limitation, a new image segmentation paradigm~\cite{yu2023convolutions,xu2023open,ding2023open,niu2024eov,qin2023freeseg} has emerged, i.e. open-vocabulary image segmentation, which enables inference over a dynamically adjustable, predefined vocabulary of categories.

\begin{figure}[t]
  \centering
  \begin{subfigure}[t]{0.48\linewidth}
    \centering
    \includegraphics[height=7cm]{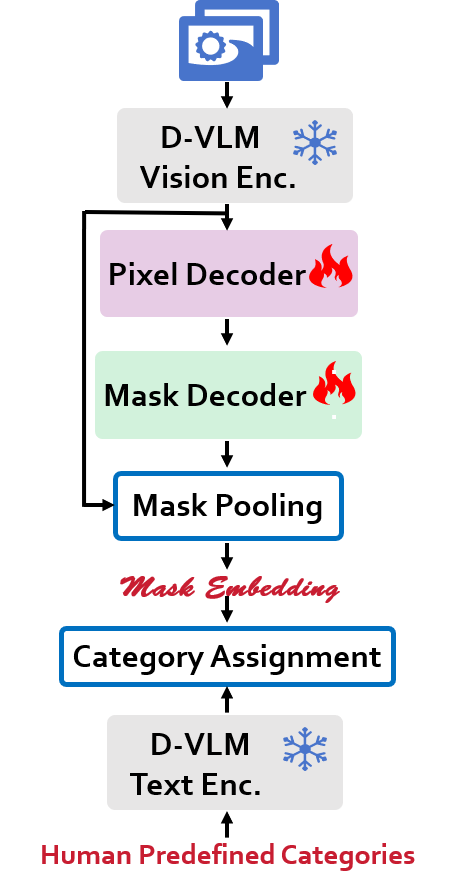}
    \caption{Human predefined open vocabulary image segmentaion.}
    \label{fig:traditional_OVIS}
  \end{subfigure}
  \hfill
  \begin{subfigure}[t]{0.48\linewidth}
    \centering
    \includegraphics[height=7cm]{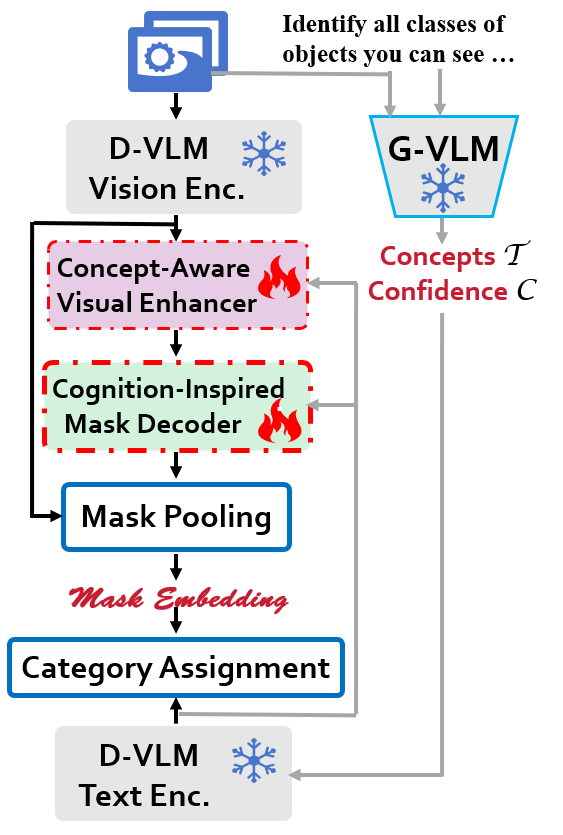}
    \caption{Cognition-inspired open vocabulary image segmentation.}
    \label{fig:our_OVIS}
  \end{subfigure}
  \caption{(a) shows traditional open vocabulary segmentation framework, which relies on human-predefined categories; (b) shows our cognition-inspired open vocabulary segmentation framework, which introduces Generative Visual-Language Model (G-VLM) to generate concepts and achieves generalized segmentation capabilities through enhancement modules.}
  \vspace{-10pt}
\end{figure}

\begin{figure}[!t]
\centering
\includegraphics[width=0.45\textwidth]{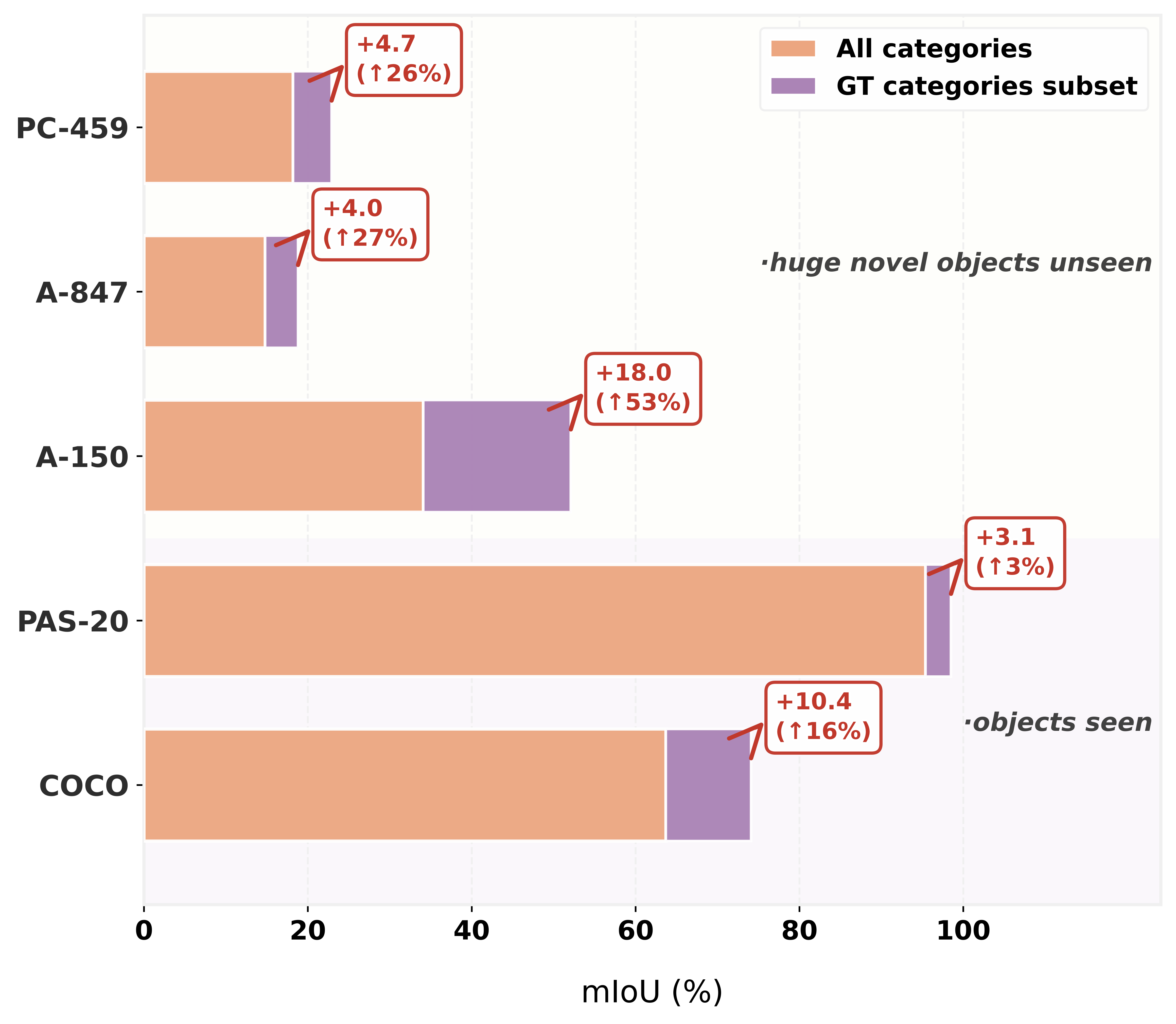}
\caption{In the traditional open vocabulary image segmentation paradigm, the mask embedding of objects is forced to approximate the few seen category embeddings in the training set, which makes it difficult for the model to accurately distinguish huge novel objects when inferring about them (All categories in the test set). We simply take a subset of the ground truth (GT) categories that only match this image for each region of the image, which can significantly improve the image segmentation performance of the model.}
\vspace{-15pt}
\label{fig:gt_classes_improvement}
\end{figure}

As illustrated in Figure~\ref{fig:traditional_OVIS}, models under this paradigm typically segment the image into multiple binary, category-independent masks. 
Each mask’s visual features, pooled from global visual representations and known as "mask embeddings", are matched with category embeddings encoded by a discriminative vision-language model (D-VLM), such as CLIP~\cite{Radford2021CLIP} or ALIGN~\cite{align}, to infer their semantic labels.
%
However, during training, mask embeddings are only aligned with a limited number of categories.
This design struggles at inference time when the model is expected to distinguish among hundreds of previously unseen categories, as shown in Figure~\ref{fig:gt_classes_improvement}. 
The root cause is a semantic disconnect—the model cannot perceive the concept it is expected to segment.

This contradicts how the human visual system operates: object recognition begins with conceptualization—the brain forms a high-level concept of the object, and then perception—the eyes delineate the object’s contours based on this concept~\cite{fenske2006top}.
%
%
%
Motivated by this cognitive process, we propose the \textbf{Cognition-Inspired Framework} for open vocabulary image segmentation (Figure~\ref{fig:our_OVIS}), which follows the human-like paradigm of Conceiving before Perceiving, encapsulated as: \textbf{What You Perceive Is What You Conceive}.
Specifically, our framework consists of three key components:
%
(1) \textbf{G-VLM Simulates Human Brain:} Traditional open vocabulary image segmentation struggle with recognizing each region of an image from hundreds of categories, despite the fact that each image typically contains only a handful of categories. This leads us to decompose the recognition task into the recognition of a subset of categories, using some prior knowledge to identify the global visual concepts in the image first, and then performing region-based recognition within this subset of concepts. Therefore, we propose to leverage a generative vision-language model (G-VLM), e.g. Qwen2.5-VL~\cite{bai2025qwen2}, Llava-NeXT~\cite{li2024llavanext-ablations}, to simulate human cognition by generating a small set of high-level object concepts for the entire image, reducing the alignment difficulty from hundreds of categories to just a few relevant ones.
(2) \textbf{Concept-Aware Visual Enhancer:} Once the object categories are identified, locating them in the image becomes analogous to referring expression segmentation~\cite{yu2018mattnet,hu2016segmentation} that locates the object according to a text description. Inspired by cross-modal fusion techniques~\cite{wang2022cris,kim2022restr}, we propose a Concept-Aware Visual Enhancer composed of: $N$-layer Text-to-Image Cross-Attention to adapt category embeddings to the current visual context, Image-to-Text Cross-Attention to guide semantic aggregation, and Deformable Self-Attention to capture structural relationships. This fusion process allows the global visual features to roughly delineate object shapes based on the generated concepts.
%
(3) \textbf{Cognition-Inspired Decoder:} The decoder integrates the generated concepts into local instance queries via a Shared Cross-Attention mechanism. These queries also extract fine-grained regional features using the same attention module. Sharing attention weights across modalities enforces modality-invariant feature learning, encouraging mask generation to be grounded in conceptual understanding—even for novel categories.
%
Given the current limitations of G-VLMs (e.g., imperfect recognition, as shown in Figure~\ref{fig:llvm}), we introduce a confidence correction strategy. During inference, we compute token-level confidence scores for each generated concept by averaging the token probabilities associated with that concept. These scores are then used to adjust the final category prediction for each instance.

%
Extensive experiments validate the effectiveness of our framework. Using only COCO Panoptic training data, our model achieves $27.2$ PQ, $17.0$ mAP, and $35.3$ mIoU on A-150, as well as $44.1$ PQ, $26.5$ mAP, and $56.2$ mIoU on Cityscapes. It further achieves $18.2$ PQ and $28.2$ mIoU on Mapillary Vistas, and $15.4$, $59.2$, $18.7$, $81.8$, and $95.8$ mIoU on A-847, PC-59, PC-459, PAS-21, and PAS-20, respectively.
This shows that our framework can effectively generalize not only to scenarios with similar distribution to the training set, but also to autonomous driving scenarios with significantly different distributions, as well as to scenarios of a large number of categories.
%
Moreover, our framework supports vocabulary-free segmentation, eliminating the need for human-defined categories. In this mode, it achieves $11.6$, $10.5$, $32.7$, $48.6$, and $95.3$ mIoU on A-847, PC-459, A-150, PC-59, and PAS-20, respectively, showcasing its versatility and generalization ability across diverse scenarios.

%% file: sec/2_related_work.tex
\section{Related Work}
\label{sec:related_work}
\subsection{Open Vocabulary Image Segmentation}
\label{sec:related_opis}
%
Open Vocabulary Image Segmentation (OVIS) represents a significant advancement in the field of image segmentation, aiming to segment and recognize objects beyond a fixed set of predefined categories. 
This task leverages the capabilities of vision-language models (VLMs) to associate visual regions with arbitrary textual descriptions, thereby enabling the identification of novel object categories.
%
Early approaches predominantly adopted a two-stage framework~\cite{xu2021simbaseline,ding2023open,liang2023ovseg,qin2023freeseg,chen2023open}. In this paradigm, the first stage involves generating class-agnostic mask proposals, while the second stage classifies these masks by matching them with textual embeddings derived from VLMs such as CLIP~\cite{Radford2021CLIP} and ALIGN~\cite{align}. While effective, this decoupled approach often entails multiple feature extraction and processing stages, leading to increased computational complexity and potential inefficiencies.
%
%
For instance, MaskCLIP~\cite{ding2023open} integrates learnable mask tokens with CLIP embeddings and class-agnostic masks to enhance segmentation performance.
OPSNet~\cite{chen2023open} combines query embeddings with the final-layer CLIP embeddings and utilizes an IoU branch to filter out less informative proposals. 
FreeSeg~\cite{qin2023freeseg} optimizes a unified network through one-time training, the same architecture and inference parameters are used to handle multiple image segmentation tasks, and adaptive cue learning is introduced to improve the robustness of the model in multiple tasks and different scenarios.
In contrast, one-stage framework~\cite{xu2023open,yu2023convolutions,niu2024eov,li2024omg} aim to unify mask generation and classification into a single, streamlined process. This approach enables a model to simultaneously generate masks and predict their corresponding categories, integrating segmentation and recognition tasks.
For example, ODISE\cite{xu2023open} employs a text-to-image diffusion model to generate mask proposals and perform classification. 
FC-CLIP\cite{yu2023convolutions} proposes a single-stage framework for segmentation leveraging a frozen convolution-based CLIP backbone.
EOV-Seg~\cite{niu2024eov} significantly improves the segmentation efficiency and performance of the model in open vocabulary scenarios through a single-stage, shared, efficient and spatially aware design. 
%
Despite the advancements in both two-stage and one-stage frameworks, these methods often diverge from human cognitive processes and encounter limitations in recognizing unseen categories. Our approach seeks to emulate human cognition by enhancing the model's perceptual capabilities, thereby achieving superior segmentation performance and more human-like understanding of novel objects.
\subsection{Large Vision-Language Model}
\label{sec:related_lvlm}
%
%
Early VLMs were primarily discriminative (D-VLM), such as CLIP~\cite{radford2021learning} and ALIGN~\cite{jia2021scaling}, employing contrastive learning to capture shared semantic features between images and texts. These models facilitate robust multimodal understanding by aligning visual and textual representations.
%
Subsequent advancements led to the emergence of generative Vision-Language Models (G-VLMs). 
For instance, BLIP~\cite{li2022blip} utilizes an autoregressive generation approach to jointly process image and text inputs, enabling tasks like image captioning and visual question answering. 
LLaVA~\cite{liu2023visual} and Qwen-VL~\cite{bai2023qwen} adopted more efficient and powerful architectures, and are trained on extensive datasets.
LLaVA extracts visual features and feeds them into large language models (LLMs) via a projection layer for end-to-end multimodal reasoning, while Qwen-VL incorporates techniques like naive dynamic resolution and multimodal rotary position encoding (M-RoPE) to improve spatiotemporal modeling for high-resolution images and videos.
In the context of open vocabulary image segmentation, D-VLM leverage their robust semantic alignment capabilities by matching image regions with a predefined vocabulary, thereby achieving precise identification and segmentation of various objects~\cite{jiao2024collaborative,niu2024eov,liang2023ovseg,qin2023freeseg,chen2023open,xu2023open,yu2023convolutions}.
Conversely, G-VLM based methods facilitate vocabulary-free image segmentation by generating descriptive text to assist the segmentation process, offering detailed segmentation cues and contextual information for improved recognition of novel objects~\cite{rewatbowornwong2023zero,reichard2025open,ulger2023auto,kawano2024tag}.
%
Distinct from these approaches, our framework aims to harness the powerful understanding capabilities of generative VLMs to augment traditional discriminative VLM-based methods. By integrating generative insights, we strive to enhance the semantic comprehension of novel objects, aligning the segmentation process more closely with human object recognition and enabling vocabulary-free image segmentation.

%% file: sec/3_method.tex
\section{Method}
\label{sec:method}
\input{figures/tex/framework_tex}
\subsection{Predefined Vocabulary in Segmentation}
%
Open-vocabulary image segmentation encompasses panoptic, instance, and semantic segmentation.
%
Panoptic segmentation simultaneously performs semantic and instance segmentation, uniformly handling both countable objects (thing classes) and uncountable backgrounds (stuff classes).
Given an input image $\mathbf{I}$ and an open vocabulary set $C_{open}=C_{train} \cup C_{test}$ comprising both thing and stuff categories, where $C_{test}$ may include novel classes not present in $C_{train}$, the objective of open vocabulary panoptic segmentation is to generate panoptic mask $\mathbf{P} = {(m_i, \hat{c}_i)}_{i=1}^N$. Here, each $m_i$ denotes a non-overlapping region (mask) in the image, and $\hat{c}_i \in C_{open}$ represents its corresponding semantic label.
Although the task is not constrained to a fixed vocabulary, existing paradigms still require a manually predefined vocabulary. Typically, these methods first segment regions in a category-agnostic manner, and subsequently assign category labels by matching visual features with text embeddings generated by a frozen, aligned, discriminative visual-language model (D-VLM).
\subsection{Cognition-Inspired Framework}
\subsubsection{G-VLM Simulates Human Brain}
\label{method:sim_brain}
%
Traditional image segmentation paradigms operate in a way that contrasts with how the human visual system recognizes objects. In the human brain, object recognition typically begins with a conceptual understanding of the object, followed by directing visual attention to delineate its contours—ultimately completing the recognition process~\cite{fenske2006top}.
%
This fundamental difference limits the ability of traditional methods to classify image regions from  hundreds of novel categories, as illustrated in Figure~\ref{fig:gt_classes_improvement}.
%
However, most images contain only a limited number of distinct visual concepts. This observation motivates the use of a generative visual-language model (G-VLM) to provide a global visual concept of the image. 
Consequently, the segmentation process only needs to match regions against a small, conceptually relevant subset of categories.
%

Specifically, given an image $\mathbf{I}$ and a simple prompt $\mathbf{P}$ “Identify all nonredundant classes of objects you can see …”, the G-VLM generates a set of visual concepts $\mathcal{T}$ with associated confidence scores $\mathcal{C}$, where each confidence score represents the average probability of the tokens comprising each concept: 
\begin{equation}
    (\mathcal{T}_{i},\mathcal{C}_{i})_{i=1}^N = G-VLM(\mathbf{I},\mathbf{P}).
\end{equation}
We then encode the text-based concepts using a pre-trained, aligned discriminative visual-language model (D-VLM, e.g., CLIP~\cite{Radford2021CLIP}) to produce semantic clues for the Concept-Aware Visual Enhancer (Section~\ref{method:cave}) and Cognition-Inspired Decoder (Section~\ref{method:cid}).
%
It is important to note that current G-VLMs are far from reaching human-level concept recognition. The focus of this work is not to improve G-VLM capabilities per se, but to explore an open-vocabulary segmentation framework inspired by the human object recognition process. Improving the capabilities of G-VLM is left as future work.
\subsubsection{Concept-Aware Visual Enhancer}
\label{method:cave}
In referring expression segmentation (RES) tasks~\cite{yu2018mattnet, hu2016segmentation}, the object to be segmented is described using a human-defined textual expression. Most existing methods~\cite{wang2022cris, kim2022restr} fuse the textual description with global visual features of the image, enabling the model to roughly localize the object based on the semantics of the expression.
%
In contrast, traditional open-vocabulary segmentation paradigms rely on a manually predefined set of category descriptions encompassing all categories in the training or test set. During inference, many of these categories are irrelevant to the content of a given image. As a result, there is little effort in prior work to integrate category descriptions with global visual features.

Leveraging G-VLM, we can first identify the relevant concepts $\mathcal{T}$ present in an image. At this point, the segmentation problem becomes similar to the RES task but uses only category-level descriptions. Inspired by this, we propose the Concept-Aware Visual Enhancer (CAVE) to fuse the identified concepts $\mathcal{T}$ with the image’s global visual features $\mathcal{V}_g$.
%
Specifically, as illustrated in the purple module of Figure~\ref{fig:framework}, CAVE comprises $N$ stacked layers, including Text2Image Cross-Attention (T2I), Text FFN (TF), Image2Text Cross-Attention (I2T), Image FFN (IF), and Deformable Self-Attention (DSA).
%
In T2I, the category embeddings $\mathcal{E}_c$ (encoded from $\mathcal{T}$ via the D-VLM text encoder) serve as the query, while the global visual features $\mathcal{V}_g$ serve as the key and value. This cross-attention allows the category embeddings to perceive image content. The transformed features are then passed through TF to yield updated concept embeddings $\mathcal{E}_{cn}$.
%
Symmetrically, in I2T, $\mathcal{E}_{cn}$ acts as key and value, while $\mathcal{V}_g$ acts as the query. This allows global visual features to semantically align with the identified concepts. However, since different images may contain different categories, batched processing can lead to mismatches between visual features and category embeddings.
%
To resolve this, we introduce a mask matrix $\mathcal{M}a \in \mathbb{R}^{B \times m}$, where $\mathcal{M}a_{i,j} = 0$ if image $i$ contains category $j$, and $-\infty$ otherwise. Here, $B$ is the batch size, and $m$ is the number of categories in the batch. The semantically-aware visual features $\mathcal{V}_{a}$ are then computed as:
\begin{equation}
 \mathcal{V}_{a} = \text{Softmax}\left( \frac{\mathcal{V}_gW_q \cdot \mathcal{E}_{cn}W_k^\top}{\sqrt{dim}} + \mathcal{M}a \right) \mathcal{E}_{cn}W_v,
\label{equ:masked-ca}
\end{equation}
where $W_q$, $W_k$ and $W_q$ are learnable projection matrices, and dim is the projection dimension.
Subsequently, the Image FFN (IF) further refines $\mathcal{V}_{a}$ into $\mathcal{V}_{gn}$.
The Deformable Self-Attention (DSA) module then aggregates structural information from $\mathcal{V}_{gn}$ to generate the final enhanced visual representation $\mathcal{V}_{sa}$.
\subsubsection{Cognition-Inspired Decoder}
\label{method:cid}
While the Concept-Aware Visual Enhancer (CAVE) effectively strengthens global visual features, making it well suited for semantic segmentation, tasks such as instance segmentation or the "thing" part of panoptic segmentation require instance-level differentiation.
In these cases, global features alone do not provide sufficient granularity to distinguish between individual object instances.
%
To address this, we propose the Cognition-Inspired Decoder, designed to extract fine-grained discriminative features from local regions.
%
\input{tables/vocabulary-free_image_segmentation}
As illustrated in the green module of Figure~\ref{fig:framework}, this decoder consists of $M$ stacked layers of Shared Cross-Attention (SCA) and Self-Attention (SA).
%
We initialize $K$ learnable queries, which first interact with the categories embeddings $\mathcal{E}_c$ via SCA. This operation injects semantic understanding into the queries. The interaction follows a mechanism similar to that in Equation~\ref{equ:masked-ca}.
%
Subsequently, the queries attend to the semantically-enhanced visual features $\mathcal{V}_{sa}$ using the same SCA mechanism. This step enables each query to focus on spatially relevant regions in the image, providing detailed spatial awareness.
%
The use of cross-attention with shared parameters plays a key role here, it facilitates the alignment between visual and semantic features within a unified embedding space. Employing a shared parameter set across modalities promotes the learning of modality-invariant representations, helping the model bridge visual content and textual priors effectively.
%
%
Next, the self-attention (SA) layers capture contextual dependencies among the different queries, enabling better discrimination between overlapping or similar instances.
%
Finally, the $K$ queries are transformed into mask queries $\hat{q}$ through a feed-forward network (FFN), and object masks $\mathbf{M} \in \mathbb{R}^{K \times H \times W}$ are generated via inner product with the enhanced visual features: 
\begin{equation}
    \mathbf{M} = \hat{q} \cdot \mathcal{V}_{sa}.
\end{equation}
To determine semantic categories, we apply Mask Pooling to extract region-level embeddings $\mathcal{E}_m$ from the visual-language aligned global features $\mathcal{V}_g$. These are then matched against categories embeddings $\mathcal{E}_c$ to predict the category for each object:
\begin{equation}
\hat{c} = \arg\max_{C_{open}}(\text{Softmax}(\mathcal{E}_m \cdot \mathcal{E}_c)).
\label{equ:category_match}
\end{equation}
Following \cite{cheng2022masked}, we adopt binary cross-entropy loss $\mathcal{L}_{pixel}$ and dice loss $\mathcal{L}_{dice}$ for mask prediction, and cross-entropy loss $\mathcal{L}_{cls}$ for category classification. The overall training loss $\mathcal{L}$ is defined as:
\begin{equation}
    \mathcal{L} = \lambda_1 \mathcal{L}_{cls} + \lambda_2 \mathcal{L}_{pixel} + \lambda_3 \mathcal{L}_{dice},
\end{equation}
where $\lambda_1$, $\lambda_2$ and $\lambda_3$ are the hyperparameters to balance different loss.
\subsubsection{Inference}
Our framework supports two inference modes: Vocabulary-Free Mode (Ours (G-VLM Assisted)) and Open Vocabulary Mode (Ours (Predefined)).
%
In Vocabulary-Free Mode, no manual specification of inference categories is required. Instead, the model relies entirely on the visual concepts $\mathcal{T}$ generated by the G-VLM. These concepts can consist of discrete category names or descriptive phrases related to object categories. The segmentation process proceeds using only this automatically generated conceptual information.
%
Open Vocabulary Mode assumes that a predefined set of inference categories $C_{test}$ is available. Here, the object concepts $\mathcal{T}$ and their associated confidence scores $\mathcal{C}$—produced by the G-VLM—are used to reweight the final category predictions, enhancing the influence of more confidently predicted concepts.
%
To implement this, we define a weighting vector $\mathcal{W} \in \mathbb{R}^{C_{test}}$ as: 
\begin{equation}
 \mathcal{W}_i = \begin{cases} e^{\mathcal{C}_i}, & \text{if } i \in \mathcal{T}, \\
 1.0, & \text{otherwise}. \end{cases} 
\label{equ:reweight_func}
\end{equation}
Using this reweighting strategy, Equation~\ref{equ:category_match} is modified as:
\begin{equation}
\hat{c} = \arg\max_{C_{test}}(\text{Softmax}(\mathcal{W} \odot (\mathcal{E}_m \cdot \mathcal{E}_c))).
\label{equ:weighted_category_match}
\end{equation}
Additionally, because the concepts $\mathcal{T}$ predicted by the G-VLM may include labels not present in $C_{test}$, we map each concept to its nearest neighbor in $C_{test}$. This mapping is performed in the CLIP Text Encoder’s semantic embedding space to ensure compatibility with evaluation on specific benchmark datasets.

%% file: figures/tex/framework_tex.tex
\begin{figure*}[!t]
\centering
\includegraphics[width=1.0\textwidth]{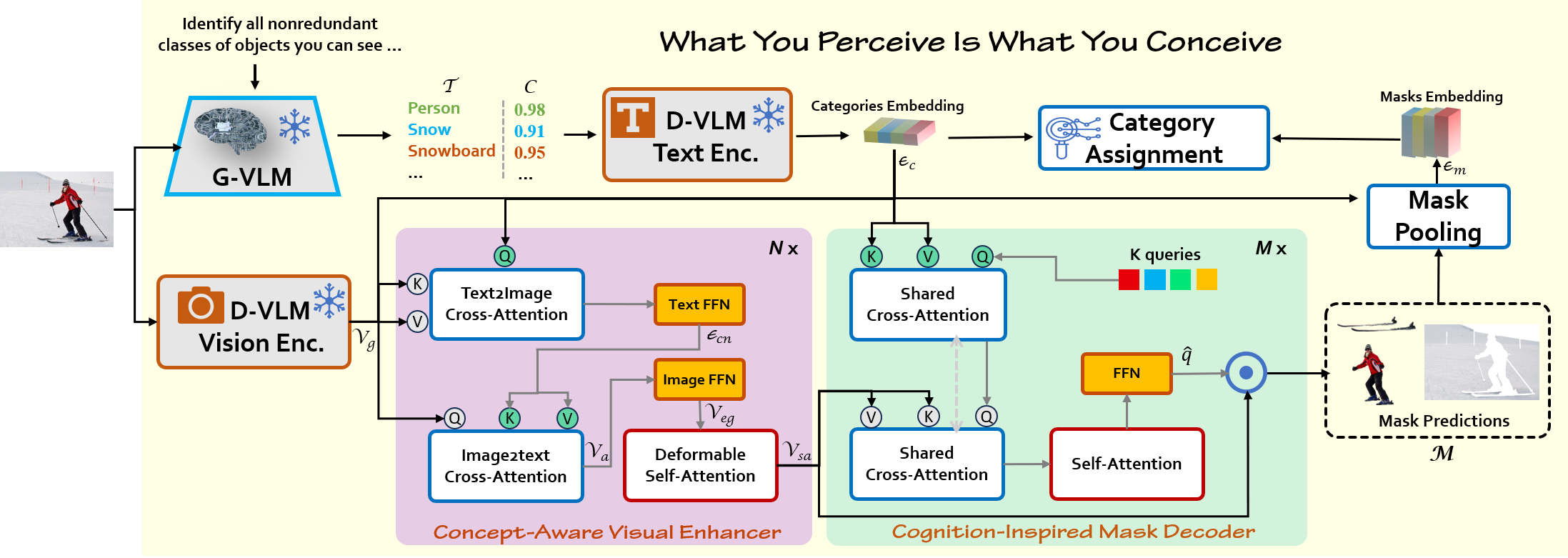}
\caption{Overview of our Cognition-Inspired Framework. This framework follows the process of human visual recognition, first Conceiving, then Perceiving, which can be summarized as What You Perceive Is What You Conceive. It first used simple prompt $\mathbf{P}$  to prompt G-VLM to generate the concepts $\mathcal{T}$  with confidence $\mathcal{C}$ . D-VLM Text Encoder encodes $\mathcal{T}$ into categories embedding $\mathcal{E}_c$. Then $\mathcal{E}_c$ interacts with the global visual features $\mathcal{V}_g$ encoded by D-VLM Vision Encoder in Concept-Aware Visual Enhancer so that $\mathcal{V}_g$ can converge according to semantic concepts. Then in Cognition-Inspired Mask Decoder, $\mathbf{K}$ queries are first fused with $\mathcal{E}_c$ to integrate semantic concepts, and then interact with the enhanced global visual features $\mathcal{V}_{eg}$ to query the objects in the image and generate the corresponding mask $\mathbf{M}$. $\mathbf{M}$ is calculated with $\mathcal{E}_c$ by cosine similarity through the local visual feature mask embedding $\mathcal{E}_m$ of Mask Pooling, and weighted by $\mathcal{C}$ to match the final category prediction.}
\label{fig:framework}
\end{figure*}

%% file: tables/vocabulary-free_image_segmentation.tex
\begin{table*}[t]
    \begin{adjustbox}{max width=\textwidth}
    \begin{tabular}{l|c|c|c|c|c|c|c|c}
    \toprule
    Method & Categories Generator & Visual Backbone & Segmentor & A-847 & PC-459 & A-150 & PC-59 & PAS-20 \\
    \midrule
    Zero-Seg~\cite{rewatbowornwong2023zero} & CLIP+GPT-2 & ViT-B/16 & Pretrained DINO~\cite{oquab2023dinov2} & - & - & - & 11.2 & 8.1 \\
    Auto-Seg~\cite{ulger2023auto} & BLIP-2 & ViT-L/16 & X-Decoder~\cite{zou2023generalized} & 5.9 & - & - & 11.7 & 87.1 \\
    TAG~\cite{kawano2024tag} & CLIP+Database & ViT-L/14 & Pretrained DINO~\cite{oquab2023dinov2} & - & - & 6.6 & 20.2 & 56.9 \\
    Chicken-and-egg~\cite{reichard2025open} & CLIP+Swin & ViT-B/16 & CAT-Seg~\cite{cho2024cat} & 6.7 & 8.0 & 18.8 & 27.8 & 81.8 \\    
    \midrule
    Ours (G-VLM assisted) & Qwen2.5-VL & ConvNeXt-L & Cognition-Inspired Mask Decoder & 11.6 & 10.5 & 32.7 & 48.6 & 95.3 \\
    Ours (Predefined) & Qwen2.5-VL+predefined Categories & ConvNeXt-L & Cognition-Inspired Mask Decoder & 15.4 & 18.7 & 35.3 & 59.2 &  95.8\\
    Ours (Ideal) & GT Categories & ConvNeXt-L & Cognition-Inspired Mask Decoder & 18.8 & 22.8 & 52.0 & 75.0 & 98.5 \\
    \bottomrule
    \end{tabular}
    \end{adjustbox}
    \caption{Comparison of vocabulary-free image segmentation methods on five benchmark datasets. The performance is reported in terms of mIoU (\%). Our framework is evaluated under three settings: G-VLM assisted (using Qwen2.5-VL for category generation), Predefined (with known test set categories, i.e. traditional open vocabulary image segmentation), and Ideal (with ground-truth categories). Our framework consistently outperforms previous methods across all datasets, demonstrating its effectiveness in vocabulary-free segmentation tasks.}
    \label{tab:vfis}
    \vspace{-15pt}
\end{table*}

%% file: sec/4_experiments.tex
\section{Experiments}
\label{sec:exp}
\subsection{Settings and Implementation Details}
\textbf{Experimental Settings.}
We evaluate our framework for vocabulary-free semantic segmentation on several popular semantic segmentation datasets: ADE20K~\cite{zhou2017ade20k} with both $150$ and $847$ class configurations, Pascal Context~\cite{6909514pascalvoc2} with $59$ and $459$ class setups, and Pascal VOC~\cite{Everingham10pascalvoc1} with its 20 classes.
Following the open vocabulary image segmentation setting~\cite{cho2024cat,xu2022simple,shan2024open}, for open vocabulary semantic, instance, and panoptic segmentation, we evaluate our framework on the COCO Panoptic~\cite{lin2014microsoft} with $133$ classes, ADE20K with both $150$ (A-150) and $847$ (A-847) class configurations, Cityscapes~\cite{Cordts2016Cityscapes} with $18$ classes and Mapillary Vistas~\cite{neuhold2017mapillary} with $65$ classes.
For open vocabulary semantic segmentation, we evaluate our framework on the PASCAL Context with $59$ (PC-59) and $459$ (PC-459) class setups and PASCAL VOC with both $20$ (PAS-20) and $21$ (PAS-21) class configurations.
For panoptic segmentation, the metrics used are panoptic quality (PQ)~\cite{kirillov2019panoptic}, Mean Average Precision (mAP), and mean intersection-over-union (mIoU), while for semantic segmentation, we use mIoU~\cite{Everingham10pascalvoc1}.

\input{tables/open_vocabulary_image_segmentation}
\textbf{Implementation Details.}
We train our model with frozen ConvNeXt-Large~\cite{liu2022convnet} visual backbone from OpenCLIP~\cite{ilharco_gabriel_2021_5143773} for $50$ epochs on COCO Panoptic~\cite{lin2014microsoft} which contains $118k$ annotated images across $133$ categories. 
Using Mask2Former~\cite{cheng2022masked} as the architecture alignment, we set the number of Concept-Aware Visual Enhancer stacks $N$ to $6$ and the number of Cognition-Inspired Mask Decoder stacks $M$ to $9$.
$\lambda_1$, $\lambda_2$, $\lambda_3$ are set to 2.0, 5.0, 5.0 respectively.
We use a crop size of $1024 \times 1024$.
The implementation and hyper-parameters setting are the same as those in Detrex~\cite{ren2023detrex}.
By default, all experiments are conducted on a single machine equipped with eight $3090$ GPUs, each with $24$ GB of memory.
For optimization, we utilize AdamW~\cite{loshchilov2017decoupled} with a learning rate $1e^{-4}$ and weight decay $0.05$.
In the evaluation, we use the same trained model, that is, the model is trained only once.
the shorted side of input images will be resized to $800$ while ensuring longer side not exceeds $1333$. For Cityscapes and Mapillary Vistas, we increase the shorter side size to $1024$.
Following prior arts~\cite{yu2023convolutions}, Mask Pooling is used to aggregate features corresponding to the mask from global visual features for category matching with the category embedding encoded by the CLIP~\cite{Radford2021CLIP} text encoder.
Unless otherwise stated, we use Qwen2.5-VL~\cite{bai2025qwen2} as our G-VLM to generate potential object concepts for an image because of its open source and powerful image understanding capabilities.
\subsection{Vocabulary-Free Image Segmentation}
In Table~\ref{tab:vfis}, we compare our framework with other state-of-the-art vocabulary-free image segmentation methods across several benchmark datasets, including A-847, PC-459, A-150, PC-59, and PAS-20.
The evaluation metric is mean Intersection-over-Union (mIoU).
Four recent vocabulary-free segmentation methods Zero-Seg~\cite{rewatbowornwong2023zero}, Auto-Seg~\cite{ulger2023auto}, TAG~\cite{kawano2024tag}, and Chicken-and-egg~\cite{reichard2025open} are compared, each of which leverages CLIP-based or BLIP-2 to generate categories and with different segmentor such as pretrained DINO, X-Decoder, or CAT-Seg. 
Zero-Seg and TAG first perform category-agnostic mask segmentation followed by category matching. As a result, they cannot perceive category information during segmenting objects, resulting in relatively low performance.
In contrast, Auto-Seg and Chicken-and-egg follow a pipeline similar to ours: object concepts are first generated and then segmented, enabling the model to leverage category information. However, their segmentors lack design elements for concept-aware perception, which limits performance.

our framework is evaluated under three category generation settings: (1) G-VLM assisted, where Qwen2.5-VL is used to generate categories; (2) Predefined, where all categories in the test set are known (traditional open vocabulary image segmentation), and Qwen2.5-VL is used to enhance category prediction weights; and (3) Ideal, where ground-truth categories in an image are used for upper-bound performance analysis.
Across all datasets, our framework consistently outperforms prior methods. In the G-VLM assisted setting, it achieves strong results (e.g., 32.7\% on A-150 and 48.6\% on PC-59), already surpassing existing methods.
Under the Predefined setting, performance improves further, reaching 35.1\% on A-150 and 59.1\% on PC-59. 
In the Ideal setting, where ground-truth categories are provided, our model achieves the best mIoU, such as 52.0\% on A-150, 75.0\% on PC-59, and 98.5\% on PAS-20, clearly demonstrating the effectiveness and scalability of our framework. 
Overall, our framework demonstrates state-of-the-art performance in vocabulary-free segmentation settings, and its upper-bound performance reveals the potential of incorporating accurate category information.
\subsection{Open Vocabulary Image Segmentation}
To further assess the generalization ability of our framework in open vocabulary image segmentation, we evaluate it on diverse benchmarks. We select A-150, PAS-20, PAS-21, and PC-59, which share similar category distributions with the COCO Panoptic training set, as well as A-847 and PC-459, which feature a larger number of categories. Additionally, we include Cityscapes and Mapillary Vistas—two autonomous driving datasets with significant domain shifts—to evaluate cross-domain robustness.
%

Table~\ref{tab:ovis} presents a comprehensive comparison of our framework against state-of-the-art methods across various tasks, including open vocabulary panoptic, instance, and semantic segmentation.
%
%
Built on the ConvNeXt-L visual backbone, our framework consistently delivers superior or highly competitive performance across all benchmarks. On the COCO Panoptic validation set, which aligns closely with the training distribution, our model demonstrates reduced overfitting compared to prior methods.
%
%
For A-150, PAS-20 and PC-59 with similar distribution, our framework yields the highest PQ, mAP and mIoU, outperforming prior methods such as ODISE~\cite{xu2023odise} and FC-CLIP~\cite{yu2023convolutions} by significant margins, thus highlighting the ability of our Concept-Aware Visual Enhancer to enhance visual features to perceive novel objects and the ability of Cognition-Inspired Decode to distinguish different instances.
Our framework also maintains strong performance in cross-domain settings. On Cityscapes and Mapillary Vistas, it achieves mIoU scores of $56.2$ and $28.2$, respectively, indicating robustness in complex real-world urban environments.
%
%
\begin{figure*}[!t]
\centering
\includegraphics[width=1.0\textwidth]{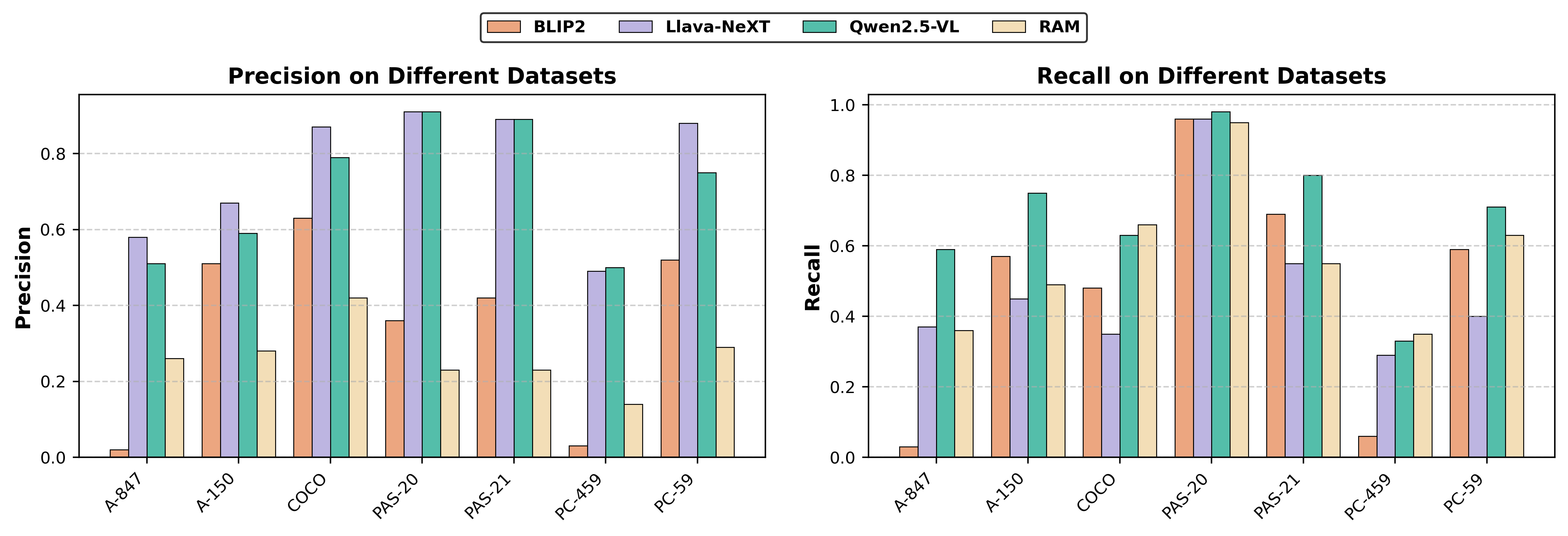}
\caption{Comparison of Precision and Recall across Different Datasets for Four Vision-Language Models (BLIP2~\cite{li2023blip}, Llava-NeXT~\cite{li2024llavanext-ablations}, Qwen2.5-VL~\cite{bai2025qwen2}, and RAM~\cite{zhang2023recognize}).}
\label{fig:llvm}
\vspace{-10pt}
\end{figure*}
On large-scale category benchmarks A-847 and PC-459, our framework continues to perform well, achieving top-tier mIoU of $15.3$ and $18.7$, respectively. This confirms its capability to generalize to diverse and fine-grained category spaces.
%
We note a performance drop on PAS-21, where our framework underperforms due to the dataset's ambiguous “background” class, which aggregates numerous stuff categories into a single label. In contrast, our model attempts to segment these into finer-grained regions, resulting in misalignments.
%
Overall, our framework achieves an excellent trade-off between performance and generalization. It either outperforms or matches state-of-the-art methods across nearly all segmentation benchmarks.
\subsection{Ablation Study}
\input{tables/abla_components}
We conduct ablation studies on the proposed key components, reweighting function defined in Equation~\ref{equ:reweight_func}, different G-VLM used to generate object concepts $\mathcal{T}$.
\noindent\textbf{Contributions of Different Components: }
Table~\ref{tab:ablation_modules} presents an ablation study on the A-150 dataset to evaluate the contribution of each component in our cognition-inspired framework. 
Starting from the baseline without any proposed modules, we observe incremental improvements by adding individual components. 
Using only the object concepts $\mathcal{T}$ generated by G-VLM without using all the categories in the test set for category matching will significantly reduce performance. This is because G-VLM is still far behind humans and cannot fully recognize what objects are in the image.
Incorporating G-VLM for generating object concepts $\mathcal{T}$ and weighting confidence $\mathcal{C}$ yields noticeable gains, particularly in mIoU$^{unseen}$. This shows that the object concept provided by G-VLM reduces the category matching from hundreds to a few, which greatly reduces the difficulty of identifying novel objects.
Adding Concept-Aware Visual Enhancer (CAVE) helps improve the global visual features, so mIoU is significantly improved.
Further adding Cognition-Inspired Decoder (CID) helps improve the distinction between instances, achieving the best results with all components enabled: $27.2$ PQ, $17.0$ mAP, $35.1$ mIoU, 48.5 mIoU$^{seen}$, and $24.7$ mIoU$^{unseen}$.
This confirms the complementary benefits of each module in improving both semantic and instance-level segmentation.
\begin{figure*}[!t]
\centering
\includegraphics[width=1.0\textwidth]{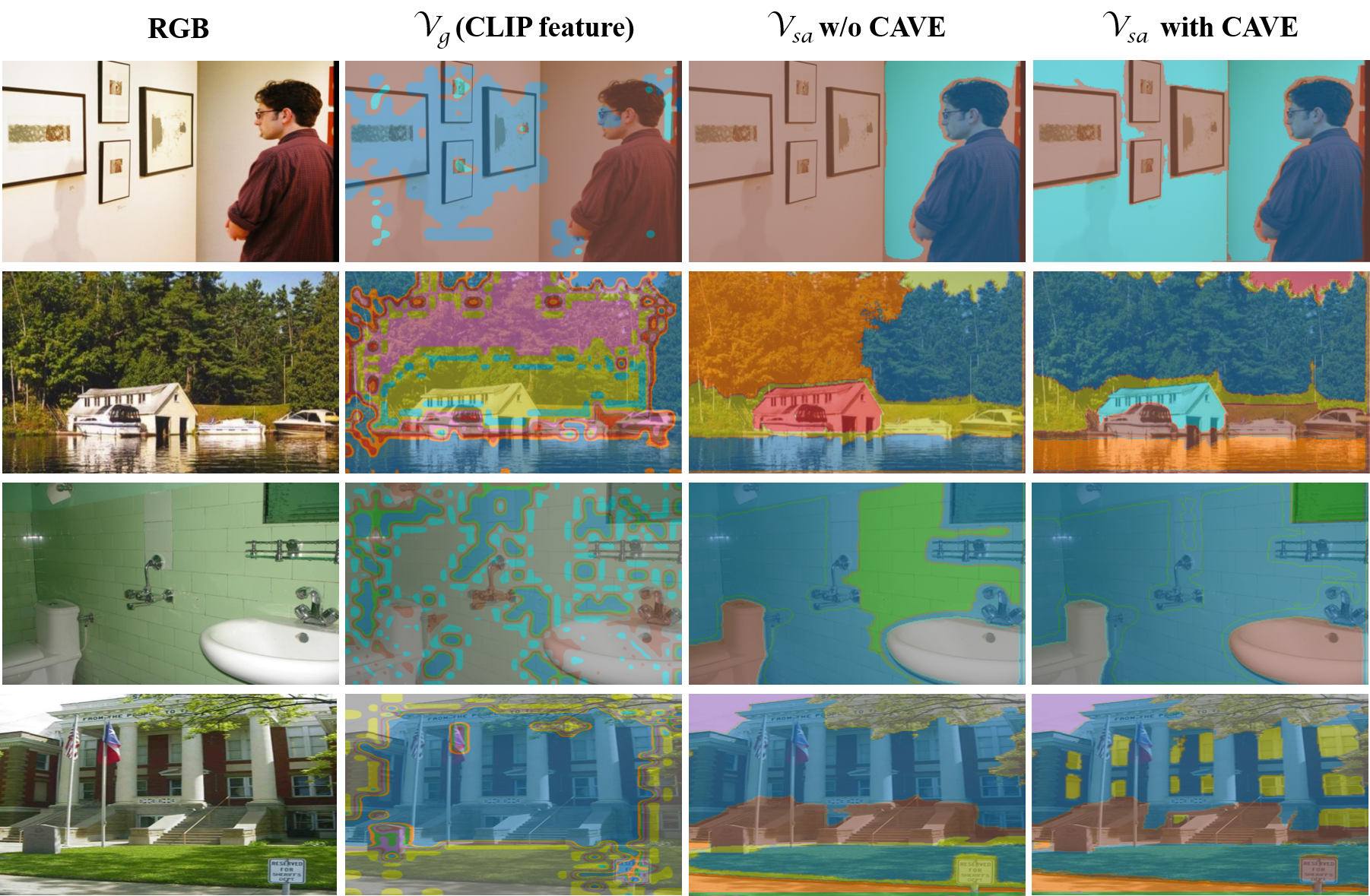}
\vspace{-15pt}
\caption{Visualization of K-means clustering of $\mathcal{V}_g$, $\mathcal{V}_{sa}$ without our Concept-Aware Visual Enhancer (replace it with Pixel Decoder~\cite{cheng2022masked}) and $\mathcal{V}_{sa}$ with our Concept-Aware Visual Enhancer.}
\label{fig:features_kmeans}
\vspace{-10pt}
\end{figure*}
\input{tables/abla_reweight_method}
\noindent\textbf{Object Confidence $\mathcal{C}$ in Class Matching Reweighting: } Table~\ref{tab:ablation_reweight} investigates the effect of different formulations of the reweighting function in Equation~\ref{equ:reweight_func} based on category confidence $\mathcal{C}$.
Evaluated across five benchmarks (A-150, A-847, PC-59, PC-459, PAS-20), all variants achieve similar performance, indicating robustness to the choice of reweighting strategy. 
The exponential formulation $\mathcal{W}_i = e^{\mathcal{C}_i}$ achieves the best overall results, particularly on PAS-20 (95.8 mIoU) and A-847 (15.4 mIoU), suggesting that exponential scaling better captures confidence distinctions for fine-grained reweighting.
\noindent\textbf{Comparison of recall and precision of different G-VLMs: } Figure~\ref{fig:llvm} compares four generative vision-language models (BLIP2~\cite{li2023blip}, Llava-NeXT~\cite{li2024llavanext-ablations}, Qwen2.5-VL~\cite{bai2025qwen2}, RAM (database based)~\cite{zhang2023recognize}) in terms of precision and recall across six benchmarks: A-847, A-150, COCO, PAS-20, PAS-21, PC-459, and PC-59.
Qwen2.5-VL consistently achieves the best balance between precision and recall, especially on COCO, PAS-20, and PAS-21, indicating its robustness across diverse visual domains.
LLaVA-NeXT shows strong precision but relatively lower recall, while RAM achieves competitive recall but suffers from low precision, particularly on A-847 and PC-459.
In addition, RAM is a way of generating categories using a database, and it cannot actually be counted as G-VLM. It is included here for analysis and comparison because Chicken-and-egg~\cite{reichard2025open} uses it to generate categories.
BLIP2 performs moderately overall, with notable recall on PAS-20, PAS-21 and PC-59. 
These results highlight Qwen2.5-VL's superior generalization in vision-language alignment for both detection accuracy and coverage.
\subsection{Qualitative Analysis}
In Figure~\ref{fig:features_kmeans}, we perform k-means clustering on the global visual features $\mathcal{V}_g$ extracted from the CLIP visual backbone and the enhanced global visual features $\mathcal{V}_{sa}$.
The first column presents the original images, followed by clustering results of the global visual features $\mathcal{V}_g$, the enhanced features $\mathcal{V}_{sa}$ obtained from the Pixel Decoder~\cite{cheng2022masked}, and those produced by our Concept-Aware Visual Enhancer.
Compared to $\mathcal{V}_g$, both versions of $\mathcal{V}_{sa}$ exhibit stronger spatial structure aggregation. However, the features enhanced by the Pixel Decoder lack semantic guidance, resulting in poor semantic convergence.
In contrast, $\mathcal{V}_{sa}$ from our Concept-Aware Visual Enhancer yields cleaner and more semantically coherent representations, characterized by clearer object boundaries and reduced fragmentation.
Notably, our enhanced features accurately capture fine-grained structures (e.g., wall frames, boats, sinks, windows in buildings) and preserve consistent region labeling even in visually complex backgrounds (e.g., foliage, tiled walls, architectural patterns).
These results underscore the effectiveness of the Concept-Aware Visual Enhancer in capturing contextual cues and maintaining semantic consistency, particularly in cluttered or low-contrast scenarios.

%% file: tables/open_vocabulary_image_segmentation.tex
\begin{table*}[t]
\begin{adjustbox}{max width=\textwidth}
\begin{tabular}{l|c|c|ccc|ccc|ccc|cc|c|c|c|c|c}
\toprule
\multirow{2}{*}{\textbf{Method}} & 
\multirow{2}{*}{\textbf{Training Dataset}} & 
\multirow{2}{*}{\textbf{Visual Backbone}} & 
\multicolumn{3}{c|}{\textbf{COCO}} & 
\multicolumn{3}{c|}{\textbf{A-150}} & 
\multicolumn{3}{c|}{\textbf{Cityscapes}} & 
\multicolumn{2}{c|}{\textbf{Mapillary Vistas}} & 
\textbf{A-847} & 
\textbf{PC-59} & 
\textbf{PC-459} & 
\textbf{PAS-21} & 
\textbf{PAS-20} \\
\cline{4-19}
& & & PQ & mAP & mIoU & PQ & mAP & mIoU & PQ & mAP & mIoU & PQ & mIoU & mIoU & mIoU & mIoU & mIoU & mIoU \\
\midrule 
ZS3Net~\cite{bucher2019zero} & Pascal VOC & ResNet101 & - & - & - & - & - & - & - & - & - & - & - & - & 19.4 & - & 38.3 & - \\
GroupViT~\cite{xu2022groupvit} & GCC~\cite{changpinyo2021conceptual}+YFCC~\cite{thomee2016yfcc100m} & ViT-S & - & - & - & - & - & 10.6 & - & - & - & - & - & 4.3 & 25.9 & 4.9 & 50.7 & 52.3 \\
OVSeg~\cite{liang2023open} & COCO-Stuff & Swin-B & - & - & - & - & - & 29.6 & - & - & - & - & - & 9.0 & 55.7 & 12.4 & - & 94.5 \\
SAN~\cite{xu2023side} & COCO-Stuff & ViT-L & - & - & - & - & - & 33.3 & - & - & - & - & - & 13.7 & 60.2 & 17.1 & - & 95.4 \\
MaskCLIP~\cite{ding2022open} & COCO Panoptic & ResNet50 & - & - & - & 15.1 & 6.0 & 23.7 & - & - & - & - & - & 8.2 & 45.9 & 10.0 & - & - \\
ODISE~\cite{xu2023odise} & COCO Panoptic & Stable Diffusion & 55.4 & 46.0 & 65.2 & 22.6 & 14.4 & 29.9 & 23.9 & - & - & 14.2 & - & 11.1 & 57.3 & 14.5 & 84.6 & - \\
MaskQCLIP~\cite{xu2023masqclip} & COCO Panoptic & ResNet50 & 48.5 & - & 62.0 & 23.3 & 12.8 & 30.4 & - & - & - & - & - & 10.7 & 57.8 & 18.2 & - & - \\
FC-CLIP*~\cite{yu2023convolutions} & COCO Panoptic & ConvNeXt-L & 56.2 & 46.9 & 65.2 & 25.3 & 16.6 & 32.9 & 43.0 & 25.1 & 53.7 & 18.3 & 27.4 & 13.8 & 56.6 & 17.9 & 82.0 & 95.2 \\ 
EOV-Seg~\cite{niu2024eov} & COCO Panoptic & ConvNeXt-L & - & - & - & 24.5 & 13.7 & 32.1 & - & - & - & - & - & 12.8 & 56.9 & 16.8 & - & 94.8 \\ 
\midrule
Ours (Predefined) & COCO Panoptic & ConvNeXt-L & 54.6 & 44.8 & 64.0 & 27.2 & 17.0 & 35.3 & 44.1 & 26.5 & 56.2 & 18.2 & 28.2 & 15.4 & 59.2 & 18.7 & 81.8 & 95.8 \\
\bottomrule
\end{tabular}
\end{adjustbox}
\caption{Quantitative comparison of segmentation performance across multiple datasets. A-150, PAS-20 and PAS-21 have a distribution similar to the training set COCO Panoptic, A-847 and PC-459 have a larger number of categories, Cityscapes and Mapillary Vistas are autonomous driving datasets with a large domain gap. PQ measures the combined quality of segmentation and recognition in panoptic segmentation, mAP evaluates instance segmentation accuracy, and mIoU reflects semantic segmentation performance. * indicates the results of retraing with their open source code.}
\vspace{-15pt}
\label{tab:ovis}
\end{table*}

%% file: tables/abla_components.tex
\begin{table}[t]
    \centering
    \begin{adjustbox}{max width=0.48\textwidth}
    \begin{tabular}{cccc|cccccc}
        \toprule
        G-VLM & Weight pred. & CAVE & CID & PQ & mAP & mIoU & mIoU$^{seen}$ & mIoU$^{unseen}$ \\
        \midrule
        \(\times\) & \(\times\) & \(\times\) & \(\times\) & 25.6 & 16.1 & 32.7 & 46.5 & 22.4 \\
        \(\checkmark\) & \(\times\) & \(\times\) & \(\times\) & 24.3 & 15.1 & 32.8 & 48.0 & 21.5 \\
        \(\checkmark\) & \(\checkmark\) & \(\times\) & \(\times\) & 26.8 & 16.7 & 34.6 & 48.7 & 24.0 \\
        \(\checkmark\) & \(\checkmark\) & \(\checkmark\) & \(\times\) & 27.0 & 16.8 & 35.0 & 49.2 & 24.3 \\
        \(\checkmark\) & \(\checkmark\) & \(\checkmark\) & \(\checkmark\) & 27.2 & 17.0 & 35.1 & 48.5 & 24.7 \\
        \bottomrule
    \end{tabular}
    \end{adjustbox}
    \caption{Ablation study on the proposed components for Our Cognition-Inspired Framework, validating on ADE20K (A-150). G-VLM: use G-VLM to generate object concepts $\mathcal{T}_{i}$, Weight pred.: use object concepts confidence $\mathcal{C}_{i}$ to correct the category prediction during inference, CAVE: Concept-Aware Visual Enhancer, CID: Cognition-Inspired Decoder.}
    \label{tab:ablation_modules}
    \vspace{-15pt}
\end{table}

%% file: tables/abla_reweight_method.tex
\begin{table}[t]
    \centering
    \begin{adjustbox}{max width=0.48\textwidth}
    \begin{tabular}{c|cccccc}
        \toprule
        $\mathcal{W}_i$ & A-150 & A-847 & PC-59 & PC-459 & PAS-20  \\
        \midrule
        $1.0 + {\mathcal{C}_i}^2$ & 35.0 & 15.1 & 59.0 & 18.6 & 95.4 \\
        \midrule
        $1.0 + \mathcal{C}_i$ & 35.1 & 15.3 & 59.1 & 18.7 & 95.5 \\
        \midrule
        $1.0 + \frac{e^{\mathcal{C}_i} - 1.0}{e - 1.0}$ & 35.0 & 15.2 & 59.0 & 18.6 & 95.5 \\
        \midrule
        $e^{\mathcal{C}_i}$ & 35.3 & 15.4 & 59.2 & 18.7 & 95.8 \\
        \bottomrule
    \end{tabular}
    \end{adjustbox}
    \caption{Ablation study on different formulations of the reweighting function $\mathcal{W}_i$ based on category confidence $\mathcal{C}_i$. The table reports performance (mIoU) across five benchmarks: A-150, A-847, PC-59, PC-459, and PAS-20.}
    \label{tab:ablation_reweight}
    \vspace{-25pt}
\end{table}

%% file: sec/5_conclusion.tex
\section{Conclusion}
\label{sec:conclusion}
We propose a Cognition-Inspired Framework significantly enhances the performance of open vocabulary image segmentation by mimicking human cognitive processes.
By first generating object concepts and then identifying their corresponding regions, the framework overcomes the limitations of conventional models in distinguish among numerous unseen categories.
%
Leveraging a G-VLM alongside the Concept-Aware Visual Enhancer and Cognition-Inspired Decoder, our framework enables more efficient and accurate segmentation.
%
Experiments on all benchmark datastets demonstrate significant performance gains.
Moreover, our framework supports vocabulary-free segmentation, improving adaptability in real-world scenarios without relying on predefined vocabulary.